\documentclass[11pt,a4paper]{article}
\usepackage[hyperref]{acl2021}
\usepackage{times}
\usepackage{latexsym}

\usepackage{microtype}

\usepackage{subcaption}

\usepackage{graphicx}
\usepackage{amsbsy}

\usepackage{amsmath}
\usepackage[ruled,vlined]{algorithm2e}
\SetInd{0.3em}{0.5em}
\SetArgSty{textup}
\aclfinalcopy 
\def\aclpaperid{630} 

\title{Human-in-the-Loop for Data Collection: a Multi-Target Counter Narrative Dataset to Fight Online Hate Speech}

\author{Margherita Fanton$^{1,2}$, Helena Bonaldi$^{1,2}$, Serra Sinem Tekiro\u{g}lu$^{2}$, Marco Guerini$^{2}$,  \\
  $^1$University of Trento, Italy \\
 $^2$Fondazione Bruno Kessler, Via Sommarive 18, Povo, Trento, Italy \\
  \texttt{mfanton@fbk.eu}, \texttt{hbonaldi@fbk.eu}, \texttt{tekiroglu@fbk.eu}, \texttt{guerini@fbk.eu}}

\date{}

\begin{document}
\maketitle
\begin{abstract}
Undermining the impact of hateful content with informed and non-aggressive responses, called counter narratives, has emerged as a possible solution for having healthier online communities. Thus, some NLP studies have started addressing the task of counter narrative generation. Although such studies have made an effort to build hate speech / counter narrative (HS/CN) datasets for neural generation, they fall short in reaching either high-quality and/or high-quantity. In this paper, we propose a novel human-in-the-loop data collection methodology in which a generative language model is refined iteratively by using its own data from the previous loops to generate new training samples that experts review and/or post-edit. Our experiments comprised several loops including dynamic variations. Results show that the methodology is scalable and facilitates diverse, novel, and cost-effective data collection. To our knowledge, the resulting dataset is the only expert-based multi-target HS/CN dataset available to the community. 
\end{abstract}

\section{Introduction}

The proliferation of online hatred  
has became an alarming issue \cite{williams2019hatred} threatening not only the well-being of target individuals and groups, but also of society as a whole. 
While authorities establish regulations and policies, social media platforms 
take actions against hate speech mostly through moderation activities, such as content removal, account suspension, or shadow-banning, at the risk of hindering the freedom of expression. Meanwhile, Non-Governmental Organizations 
are qualifying volunteers 
for responding to online hate  to promote human dignity and understanding in society. Such responses, i.e., Counter-Narratives (CN), are non-aggressive textual feedback using credible evidence, factual arguments, alternative viewpoints, and are considered as an effective strategy \cite{benesch2014countering, schieb2016governing} to confront hate speech while respecting the human rights \cite{kiritchenko2020confronting}.

However, the vast amount of online hate speech makes an effective manual intervention impossible, which motivates a line of NLP research focusing on semi or fully automatized CN generation solutions\footnote{In our view the generation  process can be fully automatic but generation systems need human supervision and should not be fully autonomous, at least for delicate tasks such as hate countering on social media platforms. For this reason we advocate that generation systems should be used as suggesting tool for NGO operators, to make their countering work more effective. In this way there is always a ``human moderator” taking the final decision \cite{conan-2019}. Furthermore, this approach is also in line  with \citet{de2017social}’s Ethical framework, since this “suggesting tool” configuration grants compliance with their rules.}. In recent years, several CN collection strategies and datasets have been proposed addressing the data-hungry nature of current state of the art generation technologies 
\cite{mathew2018analyzing,qian-etal-2019-benchmark,conan-2019}.

Considering the shortcomings of the existing collection strategies (that grant either quality or quantity, but not both), we present an approach
to produce high quality CNs for multiple hate targets while reducing the need for expert intervention. 
To this end, we build on top of the previous \textit{hybrid} data collection strategies,  
aiming to increase efficiency while maintaining the requirements of data quality, novelty and diversity. 
In particular, we start from the work by \citet{tekiroglu-etal-2020-generating} that uses an author-reviewer framework in which the author -- a generative language model -- is tasked with generating HS/CN pairs while a pool of human reviewers filter and possibly post-edit the produced output. In the present work we propose to  
further reduce the data collection effort by closing the pipeline and feeding the post-edited output back to the language model in order to regularly update it and improve the quality of the generated pairs.  
Our experiments comprised of two sessions, spanning a period of 6 months. In the \textbf{first session} we set up a `simple' human-in-the-loop (HITL henceforth) procedure and iterated it several times, measuring at each loop the performance of the whole framework according to relevant metrics. In the \textbf{second session} we run several additional loops in which we test different strategies (i.e. author configurations) to improve the data collection according to the given metrics. 
Findings show that  
the HITL framework is scalable, allowing to obtain datasets that are adequate in terms of diversity, novelty, and quantity. 
Moreover, this framework improves  
on previous hybrid data collection strategies, reducing 
at each loop the post-editing effort of the human reviewers or the number of discarded examples (session one). On the other hand, with dynamic adaptation, 
possible unwanted behaviors or flaws of the data collection can be handled at each loop by simply varying the author configuration (session 2). 
The final dataset contains 5000 HS/CN pairs in English Language, covering multiple hate targets, in terms of race, religion, country of origin, sexual orientation, disability, or gender. To the best of our knowledge, this is the first multi-target expert-based HS/CN dataset constructed through a semi-automatic mechanism and can be downloaded at the following link: \url{https://github.com/marcoguerini/CONAN}.

\section{Related Work}

With  regard  to  hatred  countering,  we  will  focus  on three research aspects relevant for the present work, i.e. (i) publicly available datasets for detection, (ii) publicly available datasets for countering, (iii) approaches for hybrid data collection.

\paragraph{Hate detection datasets.} 

Several datasets for hate detection have been presented, most of which rely on material collected from SMPs, such as Twitter \cite{waseem2016hateful, waseem2016you, ross2017measuring}, Facebook \cite{kumar2018benchmarking}, WhatsApp \cite{sprugnoli2018creating}, and forums \cite{de2018hate}. While the above datasets focus on a classification task, \citet{mathew_hatexplain_2020} released a dataset annotated with rationales to improve  hate speech interpretability and  \citet{sap-etal-2020-social} proposed the Social Bias Inference Corpus (SBIC) annotated with the description of the biases implicitly present in the language. For a more extensive review, we refer the reader to \citet{poletto_resources_2020} and \citet{vidgen_directions_2020}. 

\paragraph{Hate countering datasets.} While several social studies proved that counter-narratives are effective in hate countering \cite{benesch2014countering, silverman2016impact, schieb2016governing, stroud2018varieties, mathew2019thou}, only few works have focused on data collection for CN generation. 
\citet{mathew2018analyzing} focus on crawling, following the intuition that CNs can be found on SMPs as responses to hateful expressions. 
\citet{qian-etal-2019-benchmark} propose a crowdsourcing methodology where crowd-workers (non-expert) are instructed to write responses to hate content collected from SMPs. The study by \citet{conan-2019} also relies on outsourcing CNs writing, but via nichesourcing, using NGO operators expert in CN production. 

\paragraph{Hybrid models for data collection.} Given the data-hungry nature of current NLP technologies, one line of research has recently focused on advanced \textit{hybrid} models for data collection. 
\citet{GAN_hum_loop} proposed
using model interpretation to guide humans in the creation of adversarial examples for 
factoid question-answering systems. \citet{dinan2019build} and \citet{vidgen2020learning} perform a data collection with HITL for detecting offensive language. In both studies, the dynamic procedure is shown to be successful in reducing model error rate across rounds. \citet{vidgen2020learning} point out that the HITL approach has multiple advantages over the static data collection: design flaws can be addressed during the construction of the dataset and annotators' work is optimized, since it is guided by the feedback from the model. 
Finally \citet{tekiroglu-etal-2020-generating} propose a \textit{hybrid} approach where an LM is trained on a seed datasets of HS/CN pairs to generate new pairs that are then validated and post-edited by annotators.

\section{Methodology}

\begin{figure}[ht!]
 \centering
\includegraphics[width=.90\linewidth]{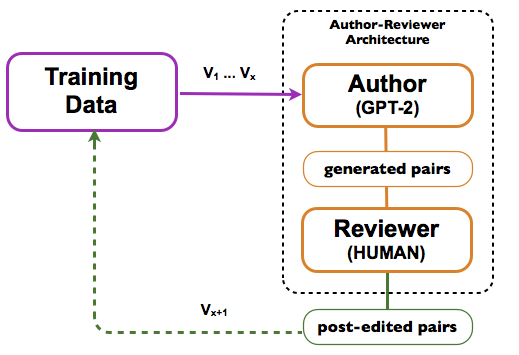}
\caption{The author-reviewer in the loop configuration. The author module produces HS/CN candidates and the reviewer(s) validates and eventually post-edits them. At each loop new examples are added to training data and the author is fine-tuned from scratch.}
 \label{fig:author-reviewer}
\end{figure}

In Figure \ref{fig:author-reviewer} we present the pipeline of our methodology. Following the idea presented by \citet{tekiroglu-etal-2020-generating}, we have an author module built using GPT-2 language model \cite{radford2019language} and fine-tuned on a seed dataset of HS/CN pairs. The author produces novel HS/CN candidates while the reviewer(s) filter and eventually post-edit them. We iterate this data collection several times, at each loop reviewed examples are added to training data and the author is fine-tuned from scratch again on all available data. In the following sections we describe the main elements used in our procedures. 

\subsection{Seed dataset}

To start the process, we built a seed dataset of 880 
HS/CN pairs by nichesourcing its collection to 20 experts from two different NGOs. We named this dataset $V_1$. The methodology for collecting $V_1$ closely replicates the one presented by \citet{conan-2019}. In particular we first created a list of prototypical hate texts -- with the help of an NGO expert -- for the following hate targets: 
\texttt{DISABLED}, \texttt{JEWS}, \texttt{OVERWEIGHT}, \texttt{LGBT+}, \texttt{MUSLIM}, \texttt{WOMEN}, 
\texttt{PEOPLE} \texttt{OF} \texttt{COLOR}, \texttt{ROMANI}, \texttt{MIGRANTS}. 
We then prepared two online data collection forms: in the first, NGO operators were asked to respond to examples selected from the prototypical hate text list, in the second they were asked to write their own HS/CN pairs.  
This data collection session lasted roughly one month.  

\subsection{Sessions}

Our experiments were run in two separate and subsequent sessions, meant to explore different aspects of the HITL approach.\\ 

\noindent In the \textbf{first session}, after using $V_1$ for the initial fine-tuning of GPT-2, we iterated the data collection 4 times, keeping the author-reviewer configuration as close as possible to the original one presented by \citet{tekiroglu-etal-2020-generating}. 
Loops are numbered sequentially as $V_2 ... V_n$. At each loop, we acquired 
500 examples of accepted and eventually post-edited HS/CN pairs\footnote{The only exception is $V_2$ that accounts for 620 pairs to have a round number of examples by reaching 1500.}.  
To obtain a new set of 500 pairs ($V_i$) we fine-tuned GPT-2 every time from scratch using 
$V_1 ... V_{i-1}$ 
as training data and administered the generated samples  
to reviewers until the target number was  
reached. In total we iterated the procedure 4 times reaching $V_5$ for a total of 3000 pairs.\\

\noindent In the \textbf{second session}, 
we tested several alternative author configurations to ameliorate some unwanted behaviors/trends that emerged during the first session. We ran 4 additional data collection loops, this time in parallel (i.e. all starting from $V_5$ dataset) instead of an iteration.  
For each loop, represented as $V_{6,\{config\_name\}}$, we collected 500 HS/CN pairs reaching a total of 5000 examples. 

\subsection{Author Models}
\thickmuskip=0mu

In our experiments all models are variants of the author (GPT-2), obtained by changing the way it is fine-tuned or conditioned. For consistency, each model is trained using the same hyperparameter configurations.   
In particular, we used GPT-2 medium model, fine-tuned for 3 epochs with a batch size of 1024 tokens and a learning rate of 2e-5. 
Each pair has been represented as $<|startofhs|> HS <|endofhs|>$ $<|startofcn|> CN <|endofcn|>$ for the training. At the generation time, Nucleus Sampling \cite{holtzman2019curious} has been utilized with a p value of 0.9.  
For the standard configurations we use only $<|startofhs|>$ for conditioning.  
Given an HS tag, the models produce a chunk of text, which is a list of HS/CN pairs. These pairs are then cleaned from the special tokens and administered to the reviewers for evaluation and possible post-editing.

\subsection{Reviewers}

We recruited 3 annotators, from a pool of internship students,
as reviewers over a period of 18 weeks to filter and post-edit 
the generated pairs after an extensive training procedure. 

\paragraph{Training.} Annotators underwent a training for 2 weeks, 
so that they became ``experts" on HS/CN post-editing.  
The training included: (i) reading and discussing NGO guidelines and public documentation describing the activity of CN writing for hate countering,  
(ii) reading all $V_1$ pairs  
to better comprehend the attributes of counter narratives, (iii) reading a sample of 100 HS/CN pairs that have been post-edited by an expert to see concrete examples of post-editing activity, (iv) performing a  
practice session of CN post-editing and discussing it with an expert NGO operator.  

\paragraph{Instructions.} We adapted the reviewing instructions from \citet{tekiroglu-etal-2020-generating}. In particular, for each pair, we asked the
operators: (a) to approve it without any modifications if it was a valid pair, (b) if the pair was not perfect, but easily amendable, to modify it, (c) if the CN is completely irrelevant, or does not follow NGO's guidelines, to discard the pair regardless of HS quality, (d) whenever there are facts or statistics in the CN, check veracity of the information to avoid possible LM hallucination effects. We further instructed the annotators to provide a hate target label for each accepted pair. 
The labels were useful both for analysis  
and for the subsequent label-based generation strategies present in $V_6$. In Table \ref{tab:post-edit-example} we give an example of GPT-2 output and its post-edited version.

\begin{table}[h!]
  \centering
  \begin{tabular}{p{0.95\linewidth}}
  \hline 
\textbf{HS:} \textit{ Transgenders should rape our children }\\
\textbf{CN:} \textit{ This is not true. Maybe they are worried because of the rise in hate crimes, incidents of which are down to 28 percent, since 2014. }\\
\hline
\textbf{HS$_{pe}$:} Transgenders want to rape our children. \\
\textbf{CN$_{pe}$:} This is not true. Maybe you should be worried about the rise in hate crimes against queers, incidents of which are almost doubled since 2014. \\
\textbf{TARGET:} \texttt{LGBT+} \\
    \hline 
  \end{tabular}
  \caption{An HS/CN example generated by GPT-2 and the post-edited version with hate target annotation.}
 \label{tab:post-edit-example}
\end{table}

\paragraph{Mitigation procedure.} 
We applied an adapted version of the guidelines by \citet{vidgen-etal-2019-challenges} to safeguard the annotators' well-being against the risk of harmful consequences of working with abusive content (present in the HSs and possibly in generated, not well-formed CNs). To this end we first made sure that annotators understood the pro-social aspects of the research and explained them the purpose of their annotation activity in details. Then we instructed the annotators to work no more than 2/3 hours per day and take regular breaks, by adjusting their workload as needed. Finally, we had meetings and feedback from the annotators on a weekly basis to let possible problems or distress emerge. This procedure was repeated throughout the whole data collection campaign.

\section{Metrics} \label{metrics}

To understand the `diachronic' behavior of our HITL methodology across iterations,  
the following metrics have been computed at the end of each loop over the newly obtained pairs. 

\paragraph{Imbalance degree} measures the difference between a perfectly-balanced distribution of the hate target categories and the actual unbalanced datasets; we use Imbalance Degree (ID) since it is specifically devoted to the multi-class scenario ~\cite{ortigosa2017measuring}. Datasets that are balanced over multiple hate targets could allow building more representative CN generation models.  

\paragraph{Acceptance Rate} is the percentage of pairs accepted by the reviewers (either untouched or post-edited) over the total number they scrutinised.  
It represents an overall estimate of the ability of the framework to produce reasonable-quality material. 

\paragraph{HTER} is originally a measure of post-editing effort at sentence level translations \cite{specia2010estimating}.  
We adopted it to the measure reviewers' effort in terms of the average number of edits over the accepted pairs. An upper-bound threshold value of 0.4 is used to account for easily post-editable pairs \cite{turchi2013coping}.

\paragraph{Novelty} measures how different  
two collections of texts are from each other, and it is grounded on Jaccard similarity.  
We utilized it to compute the originality present in $V_i$ with respect to the training data collected in previous loops \cite{dziri2018augmenting,wang2018sentigan}.

\paragraph{Repetition Rate} measures the intra-corpora quality in terms of language diversity by considering the rate of non-singleton ngram types it contains \cite{cettolo2014repetition,bertoldi2013cache}. We use it to measure the ability of the framework to provide diverse and varied examples.  
Repetition Rate (RR) has the advantage of being independent from corpus size, so it can be used to directly compare different versions of our dataset.   
 
\paragraph{Vocabulary Expansion} is a measure we introduce to serve two main objectives: (i) quantifying the contribution of the author and the reviewers, by focusing on new tokens appeared at each loop 
 (e.g. the term ``\textit{peace}" was introduced for the first time by annotators in $V_2$),  
(ii) quantifying the presence of cross-fertilization, i.e. 
tokens that appear for the first time in version $V_n$ for a particular target, but  
they were present in a version antecedent to $V_n$ for the other targets (e.g.  the term ``\textit{peace}" for the target \texttt{JEWS} appears at $V_4$ but it was already present for the target \texttt{MUSLIM} in $V_2$). The algorithm for computing Vocabulary Expansion is described in Appendix \ref{sec:VocabularyAlgorithm}.

\section{Session One}\label{SessionOne}

In session one, all the versions of the dataset $V_2...V_5$ are generated using GPT-2$_{V_i}$, where the fine-tuning is performed on all previous versions of the dataset $V_1 ... V_{i-1}$ as explained earlier. 

To produce HS/CN pairs, the author conditioning is performed using only 
\texttt{<|startofhs|>} 
tag and collecting all the generated material provided that each pair is encapsulated with the proper tags.

For the analysis, we computed the metrics described in Section \ref{metrics} on the HS/CN pairs obtained in each loop 
using micro-averaging (in Appendix \ref{AppendixTables}, Table \ref{CompleteTable} we report all results in detail).  
To isolate the possible effect of target-class imbalance, macro averages were also calculated; similarly, to account for element-wise differences we calculated micro averages for HS and CN sets separately\footnote{These results are in line with the ones showed in the paper, and do not change the discussion. They are reported in Appendix \ref{AppendixTables}, Table \ref{CompleteTableSingle}}. 

\begin{figure}[h]
\centering
\begin{subfigure}{.24\textwidth}
  \centering
  \includegraphics[width=.95\linewidth]{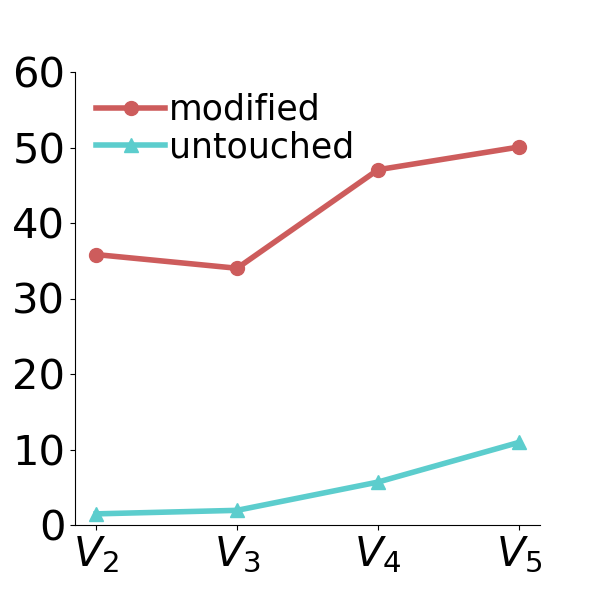}
\end{subfigure}%
\begin{subfigure}{.24\textwidth}
  \centering
  \includegraphics[width=.95\linewidth]{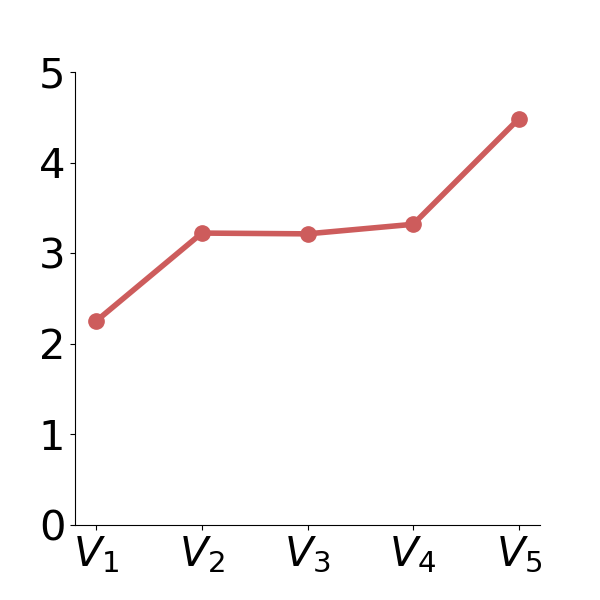}
\end{subfigure}
\caption{On the left: Percentage of pairs accepted (i) modified and (ii) untouched.
On the right: ID  
calculated over the 7 main target classes.}
\label{fig:acceptedimbalance_sp}
\end{figure}

\paragraph{Discussion.} 
Considering our objective of collecting quality material in an efficient way, we first focus on the ratio of accepted pairs and the post-editing effort in each loop. 
As shown in Figure \ref{fig:acceptedimbalance_sp}, the percentage of accepted pairs tends to increase across the loops, for both the pairs that are post-edited (``modified") from 35.8 in ${V_2}$ to 50.1 in ${V_5}$ and the ones accepted without post-editing (``untouched") from 1.5 in ${V_2}$ to 10.9 in ${V_5}$.

\begin{figure}[h]
\centering
\begin{subfigure}{.24\textwidth}
  \centering
  \includegraphics[width=.95\linewidth]{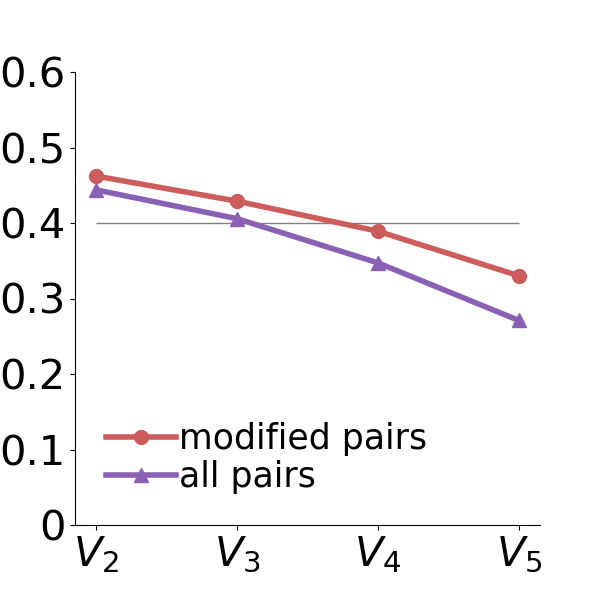}
\end{subfigure}%
\begin{subfigure}{.24\textwidth}
  \centering
  \includegraphics[width=.95\linewidth]{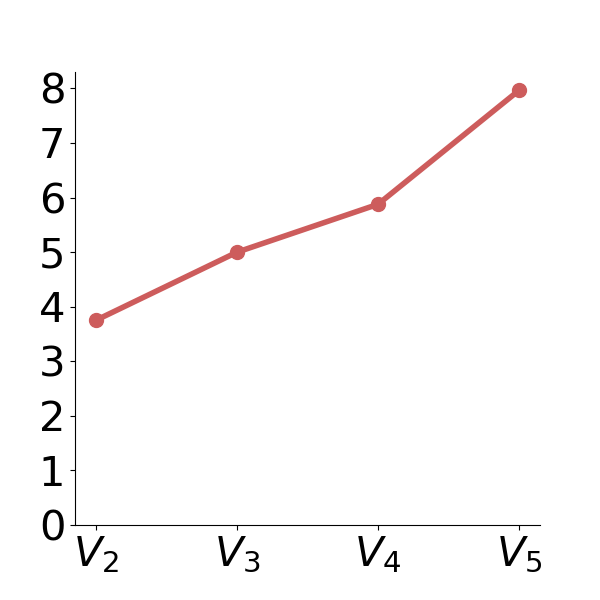}
\end{subfigure}
\caption{On the left: evolution of the post-editing effort in terms of HTER across loops both for all pairs and modified only. On the right: Micro average of Repetition Rate (RR) across loops for the HS+CN pairs. 
}
\label{fig:hterRR_session1}
\end{figure}

At the same time, the average post-editing effort of the reviewers tend to decrease across the versions, as depicted in Figure \ref{fig:hterRR_session1}. To ensure that the decrease in HTER is not due to the increasing 
ratio of untouched pairs to the total number of accepted pairs, we computed the HTER for the modified pairs alone. Consistently with the overall trend, HTER for modified pairs also declines, indicating that the data collection loops succeeded not only in reducing the reviewer effort, but also in improving the quality of the generated material to be post-edited. Notably, after $V_3$ the HTER falls  
below the 0.4 acceptability threshold as defined in \cite{turchi2013coping} for the AMT scenario (Figure \ref{fig:hterRR_session1}). In view of this analysis, we can conclude that the efficiency of data collection is increased by HITL as compared to a static approach that does not retrain the author module (that can be represented by V$_2$).

\begin{figure}[ht!]
 \centering
\includegraphics[width=.95\linewidth]{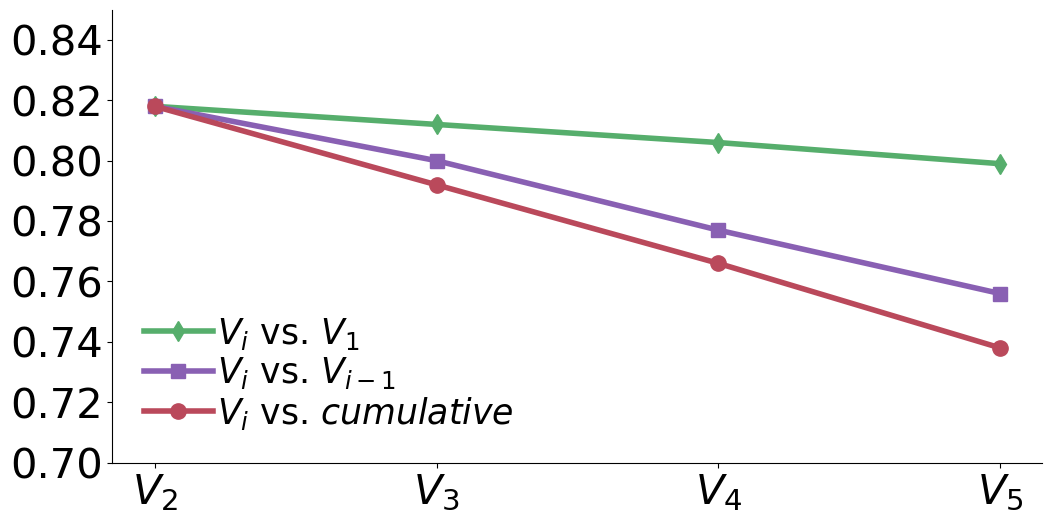}
\caption{Novelty: (i)  
${V_i}$ with respect to ${V_1}$ seed dataset,  
(ii) ${V_i}$ with respect to the previous version 
(iii) Cumulative novelty, i.e. $V_{i}$ vs. $V_1...V_{i-1}$.
}
 \label{fig:noveltymicro_sp.png}
\end{figure}

Regarding the evaluations with the quality metric Repetition Rate (Figure \ref{fig:hterRR_session1}),
it increases from $V_2$ on signifying a decrease in the lexical diversity of the generated data.
Moreover, we observed a consistent trend for the scores of the second quality metric, i.e. Novelty (Figure \ref{fig:noveltymicro_sp.png}). Similar to the diversity, novelty of the collected data also decreases across the versions, regardless of the dataset against which the novelty is computed.
Particularly, the change in the cumulative novelty represents how the vocabulary becomes less and less enrichable as the loop number increases, indicating a possible saturation point where novel material is highly difficult to obtain. Finally, the distribution of hate targets shows a worsening also in terms of ID that increases from a score of 2.2 in $V_1$ to 4.5 in $V_5$ (see Figure \ref{fig:acceptedimbalance_sp}) with some targets becoming predominant while others slowly disappearing. More details on each target distribution per loop are given in Appendix \ref{AppendixSessionOne}, Figure \ref{fig:TargetsDistribution}.

As for pair length, throughout the loops we found that ``untouched" pairs 
are usually shorter (30.7 tokens on average) than the other accepted pairs (37.3 tokens on average before post-editing). 
During the discussion sessions, annotators reported that the ``untouched" pairs are not only shorter but also somewhat stereotypical, with a small novelty added to the overall dataset (e.g. ``\textit{you cannot say this about an entire religion}", ``\textit{It's unfair to say this about an entire religion}").

\section{Session Two}\label{SessionTwo}

Given the problems emerged during the loops of the first session (i.e. higher efficiency but lower quality at each loop), we organized an additional session to test several parallel methodologies to ameliorate them. 
The description of the $V_6$ configurations are as follows:\\

\noindent$\boldsymbol{V_{6,SBF}}$ : The model GPT-2$_{V_5}$ is conditioned with novel 
offensive speeches extracted from SBIC corpus \cite{sap-etal-2020-social}. We chose this resource since: (i) it contains several thousand of  
social media posts containing biases and stereotypes spanning the same target categories with our study, (ii) for each post it provides an 
`implied statement' that closely  
resembles a `prototypical hate speech' on which we trained our system. We sampled the same number of `implied statements' for each target that maps to our labels\footnote{In Table \ref{tab:sbf_mapping} in Appendix we provide the mapping we used.} among the ones annotated with `the intent behind the statement was to offend' and/or 'the post could be offensive to someone'. We provide the statements as conditions by appending them to $<|startofhs|>$.

\noindent$\boldsymbol{V_{6,LAB}}$ : The model is conditioned specifying on which hate target it should focus on. 
In this configuration, we trained a variant of GPT-2$_{V_5}$ that takes into account the target label, and modified the original representation of our training data accordingly. In particular we accommodate hate target information within the starting token: $<|startofhs:\:target\_label|>$.\\
$\boldsymbol{V_{6,ARG}}$ : We 
fine-tuned GPT-2 on a dataset of argumentative pairs collected from Kialo\footnote{\url{www.kialo.com}}, an online debate platform for constructive and rational discussions among peers that has been exploited recently by the NLP community 
\cite{durmus2019determining,durmus2019role,scialom2020toward}. Each discussion in Kialo is represented as a tree of arguments in which a child node is connected to its parent via a ``pro" or ``con" relation.  
Extracting all the claims connected by a ``con" relation, we obtained a dataset of 128178 argument pairs covering a broader domain as compared to \texttt{HS/CN} pairs.  
We then fine-tuned GPT-2 for 1 epoch over the argumentation dataset with the standard hyperparameters. Preliminary experiments showed that the best strategy was to represent these pairs with the same format as ours to facilitate transfer of task characteristics and argumentative knowledge. 
Then this model was again fine-tuned using the standard $V_1...V_5$ data. At inference time, conditioning has been performed using lists of unique HSs from the $V_1...V_5$ data.

\noindent$\boldsymbol{V_{6,MIX}}$ : The last model is obtained by blending the three previous versions together, i.e. first fine-tuning on Kialo dataset, second fine-tuning using target label notation on $V_1...V_5$ data, conditioning using SBIC offensive speeches. \\

Bearing in mind the problems emerged during Session One, our first goal in Session Two was to balance the dataset with respect to the hate targets (i.e. reducing ID score). To this end the conditioning always takes into account the hate target label (with respect to 7 targets: 
\texttt{JEWS}, \texttt{LGBT+}, \texttt{MUSLIM}, \texttt{WOMEN}, \texttt{DISABLED},\texttt{PEOPLE} \texttt{OF} \texttt{COLOR}, \texttt{MIGRANTS}) either explicitly as in $V_{6,LAB}$ or $V_{6,MIX}$, or implicitly as in $V_{6,SBF}$ and $V_{6,ARG}$. 
In addition, to better balance the number of pairs for each target, we administered only the first 5 pairs of each generated chunk to the reviewers.

\begin{figure}[ht!]
 \centering
\includegraphics[width=.95\linewidth]{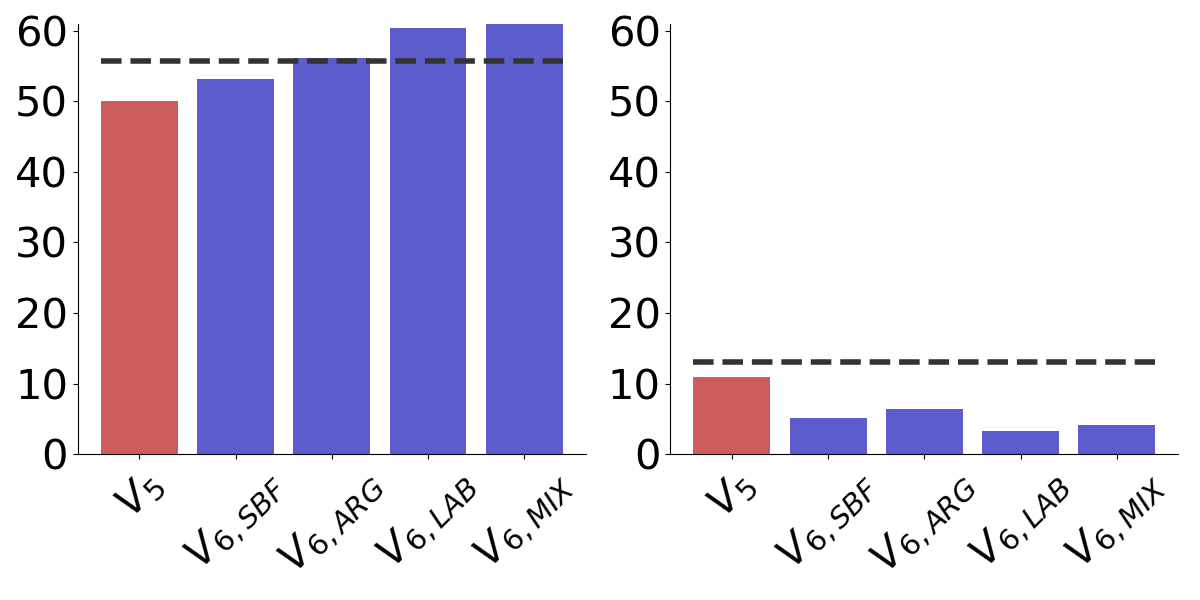}
\caption{Acceptance rate for $V_6$ configurations: modified pairs on the left, untouched pairs on the right.}
 \label{fig:acceptedversion_bars}
\end{figure}

\paragraph{Discussion.} All the applied methodologies allow for a better balancing of data in terms of hate targets, yielding an average ID score of 2.3 for the $V_6$ configurations in comparison to the ID score of 4.5 for $V_5$\footnote{In Appendix, Table \ref{tab:TargetsDistribution_dataset}, we provide the target distribution over the final dataset.}. As shown in Figure \ref{fig:acceptedversion_bars} - left, all $V_6$ configurations have a slightly higher acceptance rate than $V_5$\footnote{In order to estimate the trend of each metric after $V_5$, we calculated also $V_{6,PREDICTED}$, shown as a dashed line in the plots, using a linear regression model over $V_1...V_5$.}. 
Thus introducing novel material or data representation in 
fine-tuning stages has no strong perturbation effect. Second, and more interestingly, we observe a significant variation in the ratio of untouched and modified pairs to all the reviewed samples: for all $V_6$ approaches while there is a strong decrease in ratio of untouched pairs (Figure \ref{fig:acceptedversion_bars}, right), there is a significant increase in those modified (see Figure \ref{fig:acceptedversion_bars}, left).
In other words these models were able to produce a higher amount of suitable, albeit non perfect, pairs. In particular, comparing  ${V_6}$ configurations we can observe that for the untouched pairs the highest acceptance rate is achieved via ${V_{6,ARG}}$ with 6.37\% accepted pairs, whereas for the modified pairs 
${V_{6,MIX}}$ yields the highest percentage, with 66.15\% of the pairs accepted. 

\begin{figure}[ht!]
 \centering
\includegraphics[width=.95\linewidth]{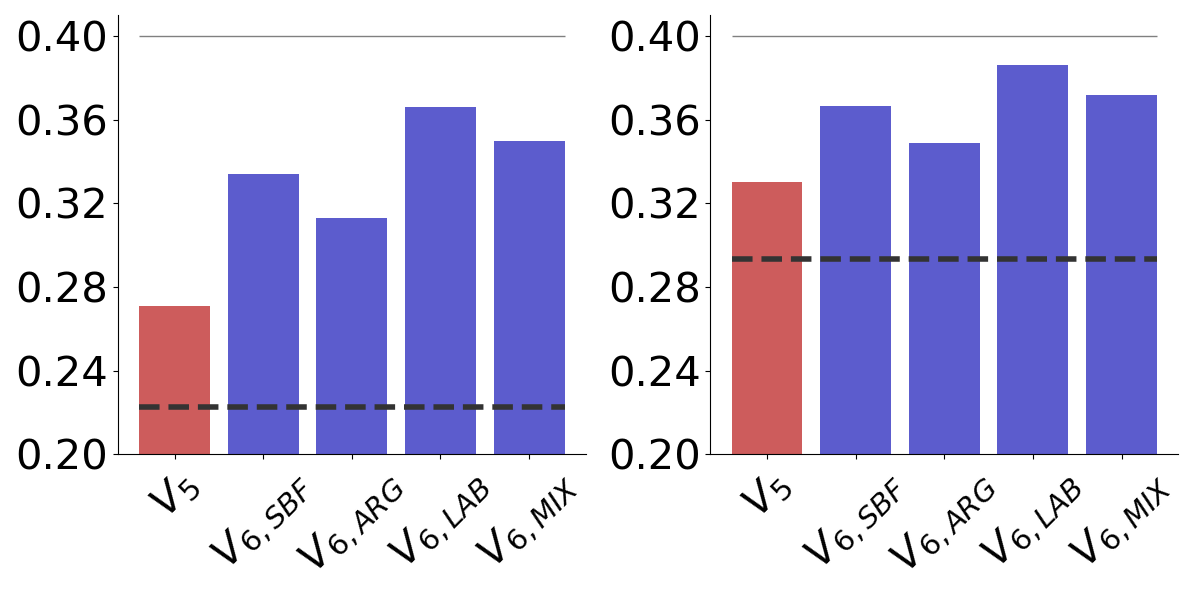}
\caption{$V_6$ configurations HTER, for all pairs on the left, modified pairs on the right.}

 \label{fig:htermicro_bars}
\end{figure}

Concerning the reviewer's effort, we see that the overall HTER increases for the all $V_6$ approaches (Figure \ref{fig:htermicro_bars}, left). Considering that we had a lower number of untouched and a higher number of modified pairs this was expected, and if we turn to the HTER of modified pairs alone we see that there is a smaller difference between $V_5$ and $V_6$ HTER. Even more interestingly, the HTER scores of all ${V_6}$ configurations, even if higher than $V_5$, are still below the acceptability threshold value of 0.4 defined earlier. Going into details, amongst the ${V_6}$ configurations, HTER reaches its lowest value in ${V_{6,ARG}}$, for both the modified and untouched pairs: 
since it was conditioned using gold HS material, this result is expected. As opposed to the other models, ${V_{6,LAB}}$ is conditioned only with a label representation and not with actual HSs.  
This affected negatively the post-editing effort, as we can notice a higher HTER for this configuration. Moreover, ${V_{6,LAB}}$ has a smaller amount of untouched pairs, so we expected HTER to spike up.

\begin{figure}[ht!]
 \centering
\includegraphics[width=.95\linewidth]{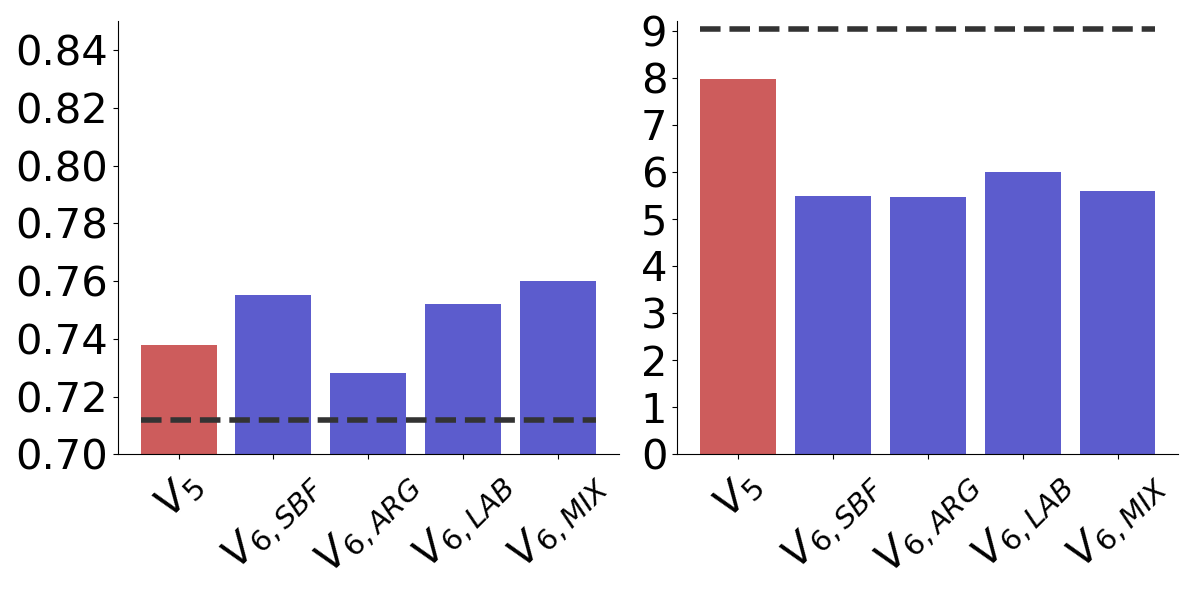}
\caption{$V_6$ configurations. Cumulative Novelty (on the left), Repetition Rate (on the right).}
 \label{fig:V6_quality}
\end{figure}

With regard to data quality (see Figure \ref{fig:V6_quality}), we see that all $V_6$ strategies succeed in increasing the novelty both with respect to ${V_5}$ and expected ${V_6}$ (the dashed line)
, except for  ${V_{6,ARG}}$, possibly due to its conditioning with HSs from ${V_1}$ ... ${V_5}$. 
Therefore, 
we also computed the novelty for CN set alone to discard the effect of HS on the metric. 
In this setting, all $V_{6}$ configurations reach a novelty between 0.741 and 0.745, as compared to a CN novelty in $V_5$ of 0.737 (as in Appendix \ref{AppendixSessionTwo}). The effect of gold HS conditioning in ${V_{6,ARG}}$ can also be spotted in the lowest HTER results in Figure \ref{fig:htermicro_bars}. The highest increase in novelty is recorded for ${V_{6,MIX}}$, reaching a score of 0.76; also novelty scores computed with respect to $V_5$ and $V_1$ confirm the result.  

All $V_{6}$ configurations succeeded in reaching an RR lower than both $V_{5}$ and expected $V_{6}$ (the dashed line). It is interesting 
that ${V_{6,LAB}}$ has the highest RR among the $V_{6}$ configurations, 
possibly because it was not built using 
any external knowledge, but 
only with a different label representation. On the other hand, ${V_{6,ARG}}$ configuration, 
for which an initial argumentation fine-tuning has been performed,  
has the lowest RR (5.474).\\

From this analysis 
we can conclude that ${V_6}$ configurations are better at producing sub-optimal material but worse at producing perfect material. Still the general quality of the pairs (in terms of novelty and RR) in Session Two is much higher than before,  
exhibiting the desired behavior for which these strategies were introduced.

\begin{figure}[ht!]
 \centering
\includegraphics[width=.8\linewidth]{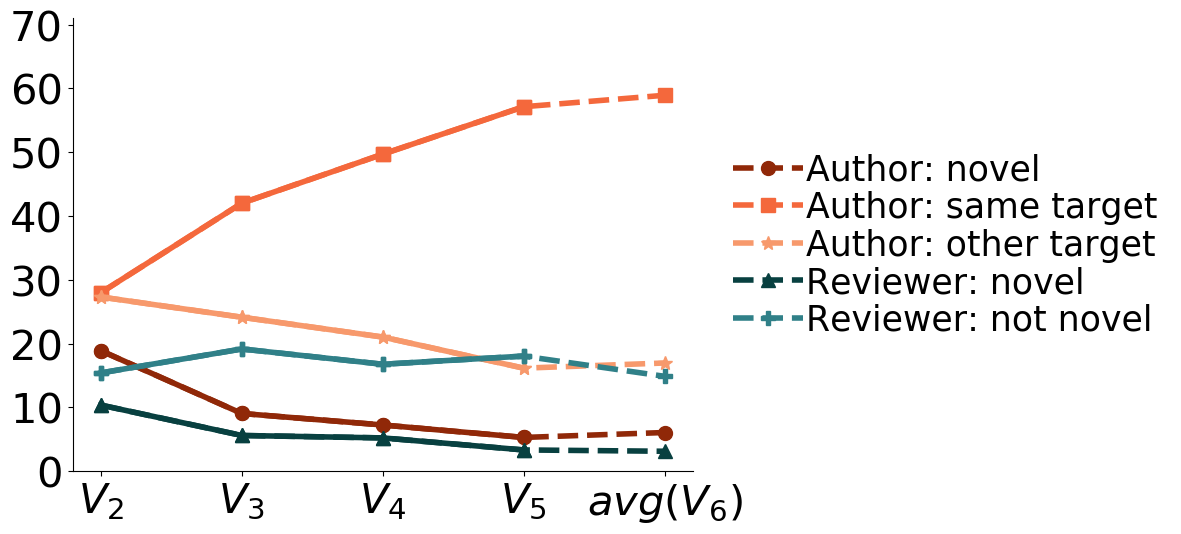}
\caption{Vocabulary expansion throughout loops (percentage of words) .}  
 \label{fig:unstackedvocAVERAGE}
\end{figure}

\section{Vocabulary analysis.} \label{vocabulary_analysis}
We report vocabulary expansion findings in Figure \ref{fig:unstackedvocAVERAGE}. For each loop $V_2 ... V_5$ the average percentage of new words injected into the dataset by the author model (GPT-2) is higher than the average percentage of new words inserted by the three reviewers during post-editing. 
Both trend-lines, even if slightly decreasing are not converging, 
implying that fine-tuned GPT-2 
is not reaching a ``saturation point” and is continuously adding new material. This trend is in line with the decrease in novelty. On the other hand, 
instructions asked for a minimal post-edit,  
so the reviewers have less opportunity to inject new material than the author and the decrease is consistent with the decreasing HTER.

As for the percentage of words generated by the author model pertaining to the same target, we see an increasing trend throughout the generations due to the cumulative nature of the metric.  
Still, the presence of words  
first observed in other targets throughout the loops, shows that the cross-fertilization phenomenon generated by GPT-2 persists. This desired feature shows the ability of GPT-2 to learn the `domain' vocabulary of other targets and to use it effectively for the target under analysis, in a way that is approved by the humans - reviewers\footnote{Even though we opted for a distinction in terms of source (same target, other target) for the `not-new' words generated by the author module, we chose not to have the same distinction for the reviewers’ `not-new' words, since we cannot assess if the reviewer was (or not) aware of the presence of a word in previous versions of the dataset.}. 
Finally, we can remark that $V_6$ versions are able to increase both the injection of novel words from GPT-2 and also its cross-fertilization ability.

\section{Qualitative Analysis}

\begin{table}[h!]
  \centering
  \begin{tabular}{p{0.95\linewidth}}
  \hline 
\textbf{HS}: \textit{``We should kill all the jews."}\\	
\textbf{CN:} \textit{``There are many alternatives to removing jews, such as converting them to another religion (e.g. Buddhism)."}\\
\hline
\textbf{HS:} \textit{``All homosexuals are trying to undermine the family by trying to impose their sexual preferences."}\\	\textbf{CN:} \textit{``Thus far, people of various sexual orientations have attempted suicide twice as often as heterosexuals. Among youth, those that seriously contemplate suicide are at almost three times the rate of heterosexual. LGB youth are almost five times as likely to have attempted suicide compared to heterosexual youth. Tell me honestly: who do you think is imposing sexual orientation to whom?"}\\
\hline
\textbf{HS:} \textit{``I don't like muggleborns, they are likely to do violence."}\\	
\textbf{CN:} \textit{``We do not say that muggleborns are less likely to commit crimes. We are saying that they are almost certainly not the case. }"\\
    \hline 
  \end{tabular}
  \caption{HS/CN examples generated by GPT-2.}
 \label{tab:qualitative-example}
\end{table}

During our exploratory experiments and the discussion sessions with the annotators, several interesting subjects have emerged, which can initiate future work. \\

\noindent\textbf{Argumentation and Counter Narratives.} 
In order to obtain even more novelty in produced pairs, ${V_{6,ARG}}$ model could be used without fine-tuning on the HS/CN dataset under the assumption that a counter argument is the same as a counter narrative. Still, the ability to argument on a variety of topics is not enough to provide a meaningful CN when prompted with an HS. 
A CN also presuppose values, so - for example - a logically \textit{valid} argument is not necessarily an \textit{acceptable} CN, 
as the first example in Table \ref{tab:qualitative-example} shows (produced by GPT-2 
fine-tuned  only on Kialo arguments).

\paragraph{New arguments or new paraphrases.} One question that emerged is whether GPT-2 is able to produce novel arguments or it is just a very sophisticated paraphrasing tool. During the discussion sessions with annotators and also by manual analysis, 
we could find CNs that contained genuinely novel arguments, which were not present in the training data but produced by GPT-2. In the 
second example in Table \ref{tab:qualitative-example}, the novel argument is about capsizing the ``imposing the homosexual agenda" argument by providing data on ``suicidal attempts among homosexual youth".

\paragraph{Novel hate targets and general knowledge.} GPT-2 proved to be able to generate HS/CN pairs also for unseen targets, including intersectional ones (e.g. ``black women"). Still the lack of a ``commonsense knowledge" can produce funny results that are beyond the scope of hallucination \cite{zellers2019defending, solaiman2019release}, such as the 
third example in Table \ref{tab:qualitative-example}, where GPT-2 addresses \textit{muggleborns} (target of hate in Harry Potter books).

\section{Conclusions}

In this paper we presented a novel HITL methodology for data collection based on an author-reviewer framework. This methodology puts together an LM and a set of human reviewers, where the LM is refined iteratively, using  data  from previous  loops that have been validated by experts. 
Experiments show that as loops are iterated, efficiency in data collection increases (acceptance rate and HTER metrics) while the dataset quality decreases in terms of novelty and diversity metrics. For this reason we experimented with additional dynamic loop adaptation that are able to increase the overall quality of the dataset without hindering the efficiency significantly.

\section*{Acknowledgments}
 
This work was partly supported by the HATEMETER project within the EU Rights, Equality and Citizenship Programme 2014-2020. We are deeply grateful to Stop Hate UK and its volunteers for their help and effort in preparing the seed dataset (version $V_1$) necessary for this work.

\bibliography{bibl}  
\bibliographystyle{acl_natbib}

\clearpage
\appendix

\section{Appendix}
\label{sec:appendix}
\setcounter{page}{1}

\subsection{Vocabulary expansion algorithm}\label{sec:VocabularyAlgorithm}
The pseudo-code for the vocabulary expansion metric described in Section \ref{metrics} can be found in Algorithm \ref{alg:vocab_expansion}.
For each version and target, we define two following sets of words:
\begin{itemize}
    \item[] $VOCAB_{pe}:$ words from the post-edited pairs 
    \item[] $VOCAB_{gen}:$ words from the generated pairs 
\end{itemize} 
A word is considered \textit{novel} when it is not present in the collective vocabulary of the previous versions: $VOCAB(V_{1,\dots,i-1})$.

\begin{algorithm}[h]
\DontPrintSemicolon

\For{each \textit{version} $V_i$}{
\For{each \textit{word w} in $V_i$}{
\eIf{\textit{w} in {\small $VOCAB_{pe}$} and \textit{w} in {\small $VOCAB_{gen}$}}{
\textit{author\_w} \textleftarrow \textit{w}\;

\eIf{\textit{author\_w} in {\small $VOCAB (V_{1,\dots,i-1})$}}{
\eIf{\textit{author\_w} in \textit{same\_target {\small $VOCAB$}}}{
{\small \textbf{\textsf{same target author\_w}}} \textleftarrow \textit{author\_w}

}{{\small \textbf{\textsf{other target author\_w}}} \textleftarrow \textit{author\_w}  \;

}}{
{\small \textbf{\textsf{novel author\_w}}} \textleftarrow \textit{author\_w} \;

}}{
\textit{reviewer\_w} \textleftarrow \textit{w}\;

\eIf{\textit{reviewer\_w} in {\small $VOCAB (V_{1, \dots,i-1})$}
}{
{\small \textbf{\textsf{not novel reviewer\_w}}} \textleftarrow \textit{reviewer\_w}\;

}{
{\small \textbf{\textsf{novel reviewer\_w}}} \textleftarrow \textit{reviewer\_w}\;
}
}
}
}
\caption{Vocabulary expansion for each target} 
\label{alg:vocab_expansion}
\end{algorithm}

Each word is assigned to one of the following sets: Author-novel, Author-same-target, Author-other-target, Reviewer-novel, Reviewer-not-novel.
Considering the size in terms of words of each set, we calculate the percentages for each target and version, so that we are able to obtain the vocabulary expansion scores as macro average percentages.

\subsection{Additional material for Session One}\label{AppendixSessionOne}

In this section, we present the most interesting results that we have obtained by analysing only the HS or the CN sets.

\begin{figure}[h!]
\centering
\begin{subfigure}{.24\textwidth}
  \centering
  \includegraphics[width=.95\linewidth]{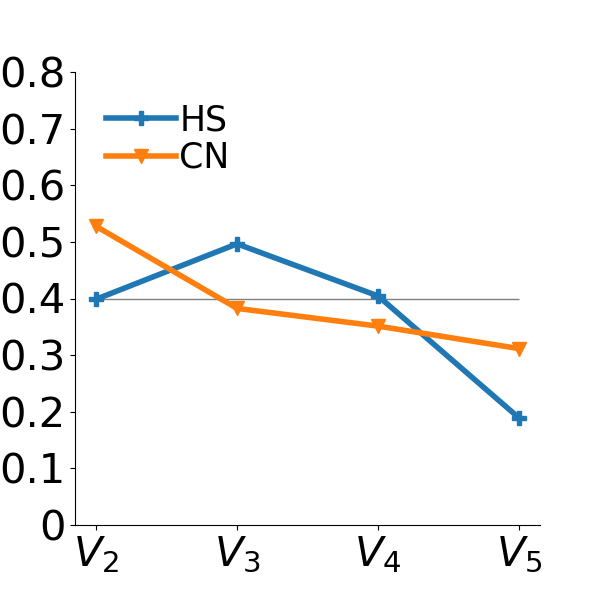}
\end{subfigure}%
\begin{subfigure}{.24\textwidth}
  \centering
  \includegraphics[width=.95\linewidth]{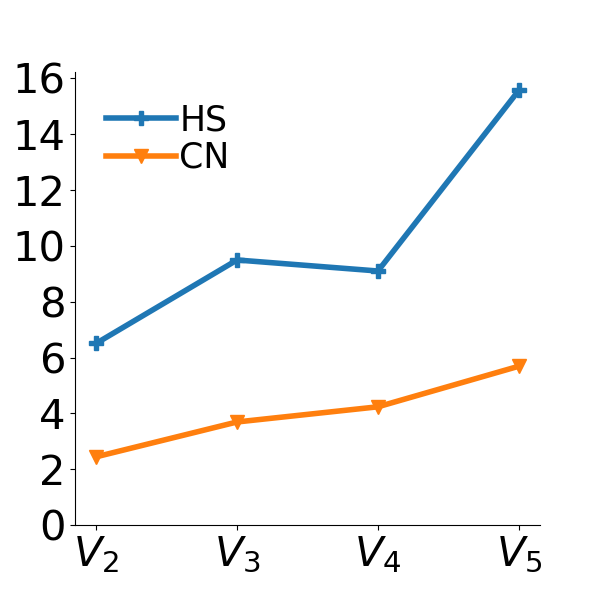}
\end{subfigure}
\caption{Session One. HTER scores on the left. RR on the rigth.}
\label{fig:HSvsCN_htermicro_RRmicro_sp}
\end{figure}

\begin{figure}[h!]
\centering
\begin{subfigure}{.24\textwidth}
  \centering
  \includegraphics[width=.95\linewidth]{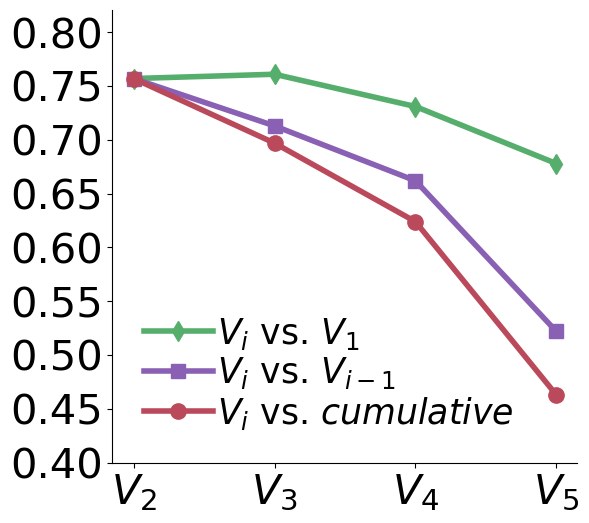}
\end{subfigure}%
\begin{subfigure}{.24\textwidth}
  \centering
  \includegraphics[width=.95\linewidth]{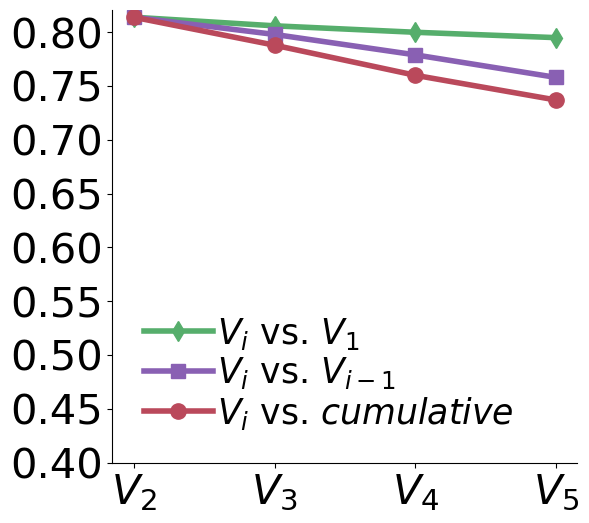}
\end{subfigure}
\caption{Session One. Novelty scores (HS on the left, CN on the right).}
\label{fig:HSvsCN_noveltymicro_sp}
\end{figure}

While HTER calculated on CN alone shows a clear decreasing trend (Figure \ref{fig:HSvsCN_htermicro_RRmicro_sp} on the left), the results for HS alone are less consistent yielding higher scores for $V_3$ and $V_4$. This can be mostly explained with the different approaches of post-editing the HSs by the annotators, which include the possibility to rewrite it entirely when needed. On the other hand, the decreasing trend of HTER for HS starting from $V_3$, resulting in a lower score in $V_5$ than the one calculated on CN only, could be due to the increasing frequency of prototypical HSs. This implication is confirmed by the higher RR scores for HSs as compared to CNs, which grow faster for the former than the latter (Figure \ref{fig:HSvsCN_htermicro_RRmicro_sp} on the right).
Moreover, the increasing number of prototypical HSs contributes to the novelty scores for HSs only being lower than those of CNs and decreasing more rapidly (Figure \ref{fig:HSvsCN_noveltymicro_sp}).  

\begin{figure}[h]
\centering
\begin{subfigure}{.24\textwidth}
  \centering
  \includegraphics[width=.95\linewidth]{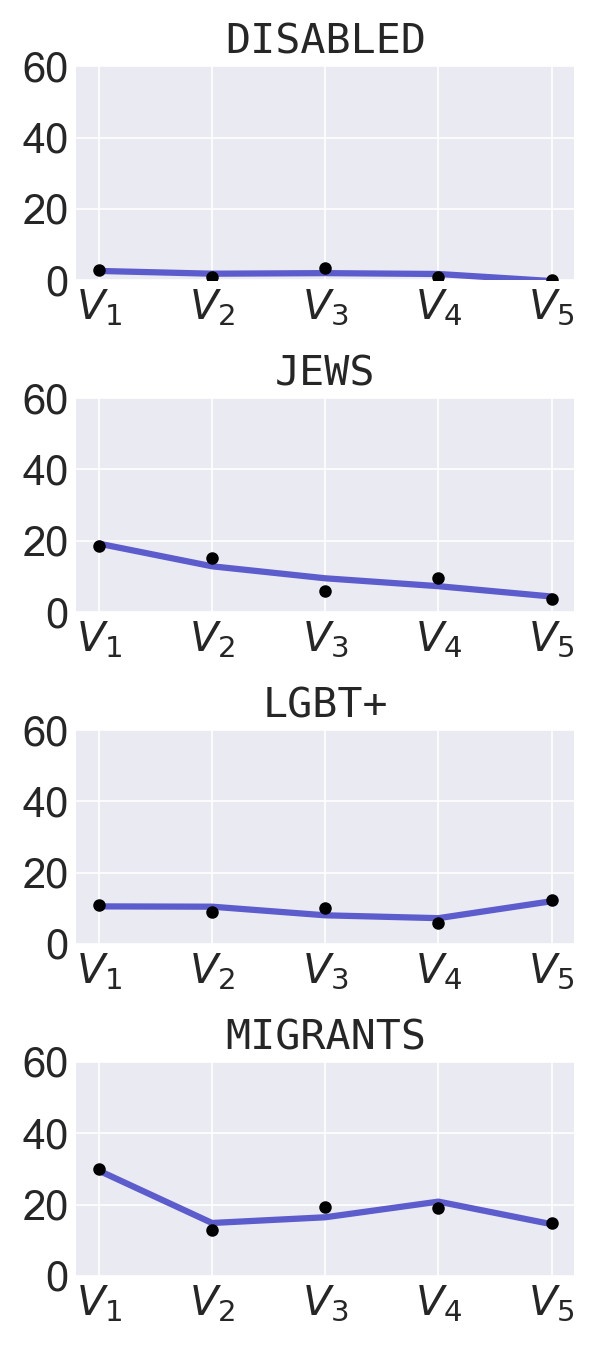}
\end{subfigure}%
\begin{subfigure}{.24\textwidth}
  \centering
  \includegraphics[width=.95\linewidth]{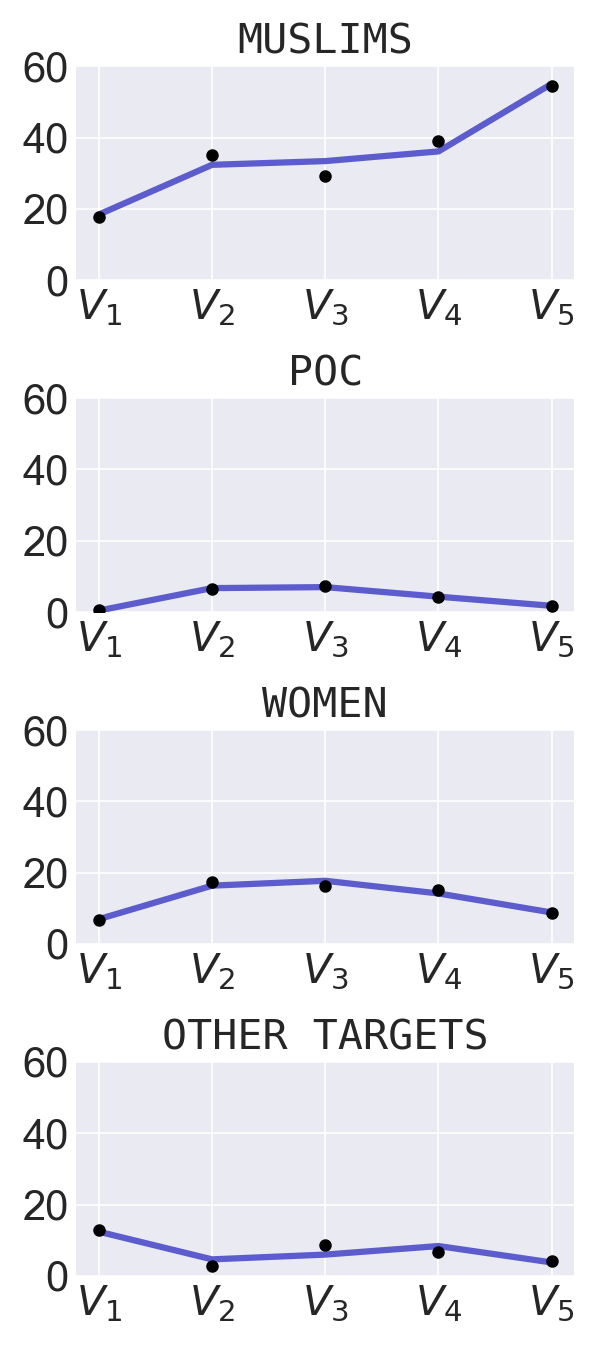}
\end{subfigure}
\caption{The targets distributions for the loops of Session One.}
\label{fig:TargetsDistribution}
\end{figure}

In Figure \ref{fig:TargetsDistribution} the target distribution at each loop of Session One is shown, in Table \ref{tab:TargetsDistribution_dataset} the  frequencies of targets in the final dataset are displayed.  The \texttt{MUSLIMS} target covers a significant percentage of the generations in every loop and consists of more than the half of the pairs ${V_5}$. In fact it is expected to cause even more imbalanced productions in the next loops. \texttt{JEWS}, \texttt{MIGRANTS} and \texttt{DISABLED} targets diminish over the loops, while the other targets can be considered as stable.

\begin{table}
\small
    \centering
    \begin{tabular}{lrr}
    \hline
         Target & Coverage & Pairs \\ \hline
        DISABLED & 4.40 & 220 \\
        JEWS & 11.87 & 594 \\ 
        LGBT+ & 12.33 & 617 \\ 
        MIGRANTS & 19.13 & 957 \\ 
        MUSLIMS & 26.68 & 1335 \\
        POC & 7.04 & 352 \\ 
        WOMEN & 13.23 & 662 \\ 
        OTHER & 5.32 & 266 \\
        \hline
        Total & 100 & 5003 \\ \hline
    \end{tabular}
     \caption{Target distribution over the final dataset.}
     \label{tab:TargetsDistribution_dataset}
\end{table}

\subsection{Additional material for Session Two}\label{AppendixSessionTwo}

\begin{figure}[h!]
 \centering
\includegraphics[width=.95\linewidth]{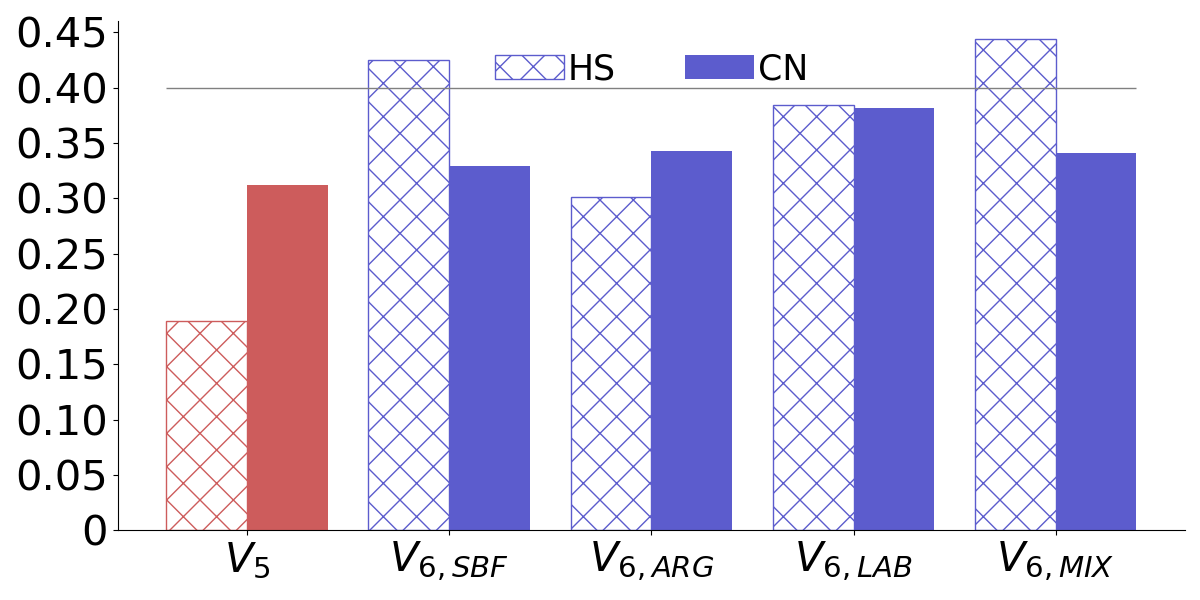}
\caption{HTER for HS and CN, computed on all pairs.}
 \label{fig:HSvsCN_htermicro_bars} 
\end{figure}

\begin{figure}[h!]
 \centering
\includegraphics[width=.95\linewidth]{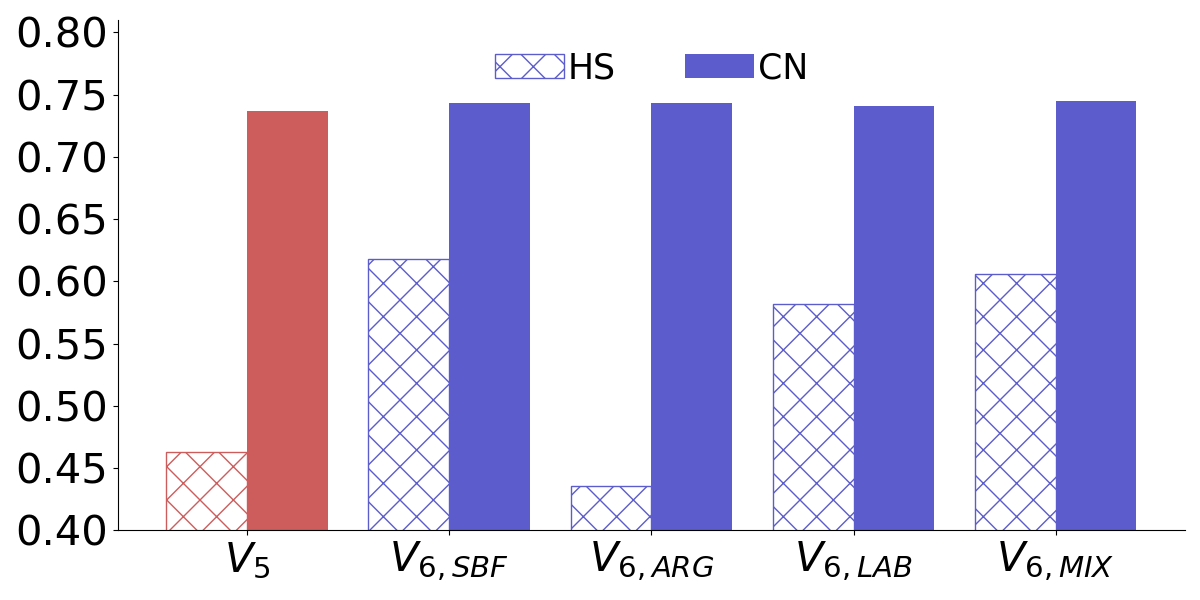}
\caption{Cumulative novelty, i.e. $V_{i}$ vs. 
$\bigcup\limits_{x=1}^{i-1} V_{x}$  for HS and CN, computed on all pairs}.
 \label{fig:HSvsCN_cumulativenoveltymicro_bars}
\end{figure}

\begin{figure}[h!]
 \centering
\includegraphics[width=.95\linewidth]{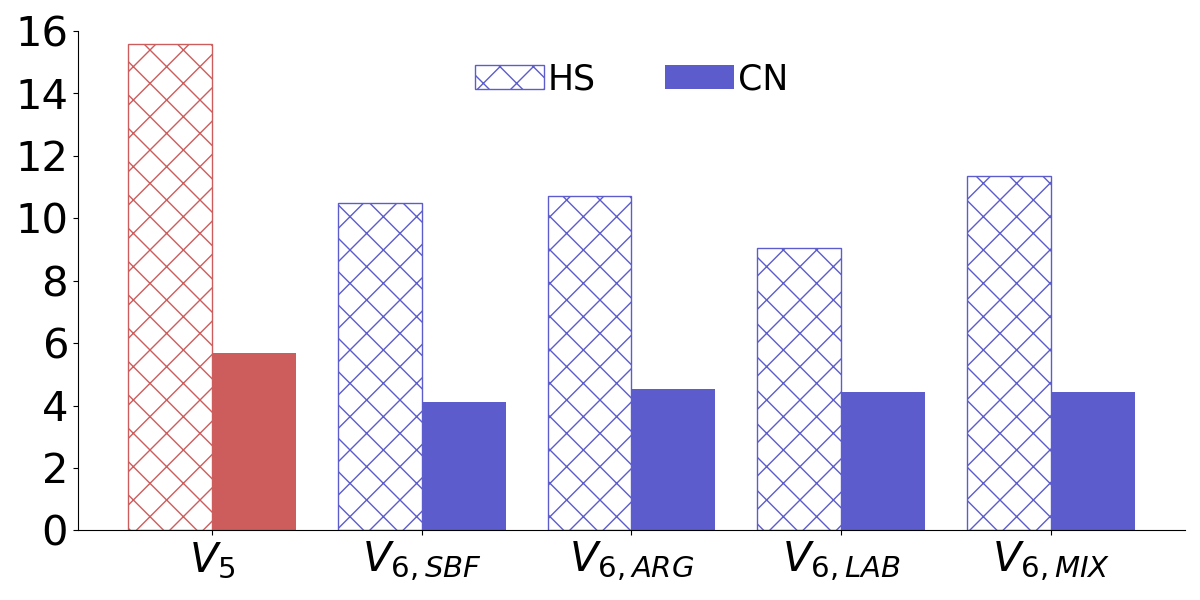}
\caption{Repetiton Rate for HS and CN, computed on all pairs.}
 \label{fig:HSvsCN_RR_bars}
\end{figure}

Concerning Session Two, the results for CNs are in line with the conclusions drawn in the paper for HS/CN pairs. The same holds for HSs, the only exception being for the cumulative novelty of $V_{6,ARG}$ HSs, as can be seen in Figure \ref{fig:HSvsCN_cumulativenoveltymicro_bars} and in Table \ref{CompleteTableSingle}. As explained earlier in Section \ref{SessionTwo}, this effect is due to the use of hate speeches from the training set for conditioning GPT-2. 
This result also corresponds to HSs from $V_{6,ARG}$ having lower HTER (Figure \ref{fig:HSvsCN_htermicro_bars}) and a higher RR (Figure \ref{fig:HSvsCN_RR_bars}). 

\begin{table}[h!]
\small
\centering
\begin{tabular}{p{0.28\linewidth}p{0.58\linewidth}}
\hline
$V_{6,SBF}$ & Labels from \citet{sap-etal-2020-social} \\
\hline
\texttt{DISABLED} & mentally disabled folks, physically disabled folks, autistic folks, blind people, folks with down syndrome, autistic \\
\texttt{JEWS} & jewish folks, jews, holocaust, holocaust victims \\
\texttt{LGBT+} & gay men, lesbian women, trans women, trans men, nonbinary folks,  gay folks, bisexual women, trans people \\
\texttt{MIGRANTS} & immigrants, illegal immigrants, refugees \\
\texttt{MUSLIM} & muslim folks, islamic folks, muslims, islamic \\
\texttt{POC} & black folks, africans, africa, people of color, african folks
 african, poc \\
\texttt{WOMEN} & women, feminists, feminist \\
\texttt{*OVERWEIGHT} & fat folks \\
\texttt{*ROMANI} & gypsies \\
\hline
\end{tabular}
\caption{Label mapping for $V_{6,SBF}$. Starred items are considered as ``other targets" in Figure \ref{fig:TargetsDistribution}.}
\label{tab:sbf_mapping}
\end{table}

\subsection{Tables}\label{AppendixTables}

In Table \ref{CompleteTable}, the main results calculated on the HS/CN pairs are displayed. In Table \ref{CompleteTableSingle}, respectively, the results calculated on HS only and CN only are shown.

\begin{table*}[hb!]
    \centering
    \resizebox{\textwidth}{!}
    {\begin{tabular}{lrrrrrrrr}
    \hline
                                   \textit{versions} & $V_{2}$ & $V_{3}$ &  $V_{4}$ &  $V_{5}$ & ${V_{6,SBF}}$ &  ${V_{6,ARG}}$ & ${V_{6,LAB}}$ & ${V_{6,MIX}}$\\
    \hline
                         imbalance degree &  3.222 &  3.214 &  3.319 &  4.485 &  2.143 &  2.095 &  3.098 &  2.057 \\
              acceptance rate (untouched) &  1.475 &  1.941 &  5.684 & 10.936 &  5.146 &  6.367 &  3.308 &  4.213 \\
               acceptance rate (modified) & 35.820 & 34.004 & 47.053 & 50.061 & 53.099 & 56.055 & 60.305 & 66.152 \\
                     discarded pairs rate & 62.705 & 64.055 & 47.263 & 39.003 & 41.755 & 37.578 & 36.387 & 29.635 \\
                         HTER (all pairs) &  0.444 &  0.406 &  0.347 &  0.271 &  0.334 &  0.313 &  0.366 &  0.350 \\
                          HTER (modified) &  0.462 &  0.429 &  0.389 &  0.330 &  0.367 &  0.349 &  0.386 &  0.372 \\
               $V_{i}$ vs. $cumulative$ novelty &  0.818 &  0.792 &  0.766 &  0.738 &  0.755 &  0.728 &  0.752 &  0.760 \\
                     ${V_i}$ vs. ${V_1}$ novelty &  0.818 &  0.812 &  0.806 &  0.799 &  0.812 &  0.795 &  0.809 &  0.813 \\
                   ${V_i}$ vs. $V_{i-1}$ novelty &  0.818 &  0.800 &  0.777 &  0.756 &  0.777 &  0.775 &  0.770 &  0.781 \\
                                       RR &  3.753 &  4.999 &  5.876 &  7.962 &  5.491 &  5.474 &  5.993 &  5.585 \\
                        vocab. GPT-2: new & 18.897 &  9.060 &  7.256 &  5.303 &  6.924 &  5.407 &  5.111 &  6.859 \\
                vocab. GPT-2: same target & 27.997 & 42.017 & 49.703 & 57.137 & 56.419 & 62.550 & 58.653 & 58.104 \\
              vocab. GPT-2: other targets & 27.316 & 24.143 & 21.039 & 16.164 & 19.353 & 14.574 & 16.607 & 17.346 \\
                        vocab. human: new & 10.373 &  5.610 &  5.230 &  3.343 &  3.040 &  3.227 &  2.844 &  3.440 \\
                    vocab. human: not new & 15.417 & 19.170 & 16.773 & 18.053 & 14.264 & 14.241 & 16.784 & 14.251 \\
    \hline
    \end{tabular}}
    \caption{All results for HS/CN pairs.}
    \label{CompleteTable}
    \end{table*}

\begin{table*}[h!]
\centering
\resizebox{\textwidth}{!}
{\begin{tabular}{lrrrrrrrr}
\hline
                     \textit{versions} &    $V_2$ &  $V_3$ &  $V_4$ &   $V_5$ & $V_{6,SBF}$ & $V_{6,ARG}$ & $V_{6,LAB}$ & $V_{6,MIX}$ \\
\hline
                     \textbf{HS metrics} &   &   &   &    & 
                     &  &  & \\
\hline
                       HTER (all pairs) &   0.399 &  0.497 &  0.405 &   0.189 &       0.424 &       0.301 &       0.385 &       0.444 \\
                         RR &  6.508 &  9.496 &  9.101 &  15.576 &       9.062 &      10.479 &        10.700 &      11.361 \\
 $V_{i}$ vs. $cumulative$ novelty &  0.757 &  0.697 &  0.624 &   0.463 &       0.618 &       0.436 &       0.582 &       0.606 \\
       ${V_i}$ vs. ${V_1}$ novelty &  0.757 &  0.761 &  0.731 &   0.678 &        0.760 &       0.689 &       0.743 &       0.758 \\
     ${V_i}$ vs. $V_{i-1}$ novelty &  0.757 &  0.713 &  0.662 &   0.522 &       0.673 &       0.652 &       0.645 &       0.666 \\
\hline
                     \textbf{CN metrics}  &   &   &   &    & 
                     &  &  & \\
\hline
                       HTER (all pairs) &  0.527 &  0.383 &  0.351 &  0.312 &       0.329 &       0.342 &       0.382 &       0.341 \\
                         RR &  2.443 &  3.692 &  4.236 &   5.690 &       4.428 &       4.125 &       4.521 &       4.428 \\
 $V_{i}$ vs. $cumulative$ novelty &   0.814 &  0.788 &   0.760 &  0.737 &       0.743 &       0.743 &       0.741 &       0.745 \\
       ${V_i}$ vs. ${V_1}$ novelty &   0.814 &  0.806 &    0.800 &  0.795 &       0.805 &       0.801 &       0.801 &       0.802 \\
     ${V_i}$ vs. $V_{i-1}$ novelty &  0.814 &  0.798 &  0.779 &  0.758 &       0.771 &       0.771 &       0.768 &       0.774 \\
\hline
\end{tabular}}
\caption{All results for HS and CN sets alone.}
\label{CompleteTableSingle}
\end{table*}

\begin{table*}[h!]
  \centering
  \begin{tabular}{p{0.95\linewidth}}
    \hline
\textbf{HS:} \textit{ Every Muslim is a potential terrorist, even if they are not born here! }\\
\textbf{CN:} \textit{ A person's place of birth is important. Some people prefer to identify as non-Muslim, some as Muslim, and still others as Christian. I prefer to call them what they are: people. }\\
\textbf{HS$_{pe}$:}  Every Muslim is a potential terrorist, even if they are born here! \\
\textbf{CN$_{pe}$:} A person's place of birth is not relevant, but neither is a person's religion. Some people are Muslims. I prefer to call them what they are: people, not potential terrorists. \\
\textbf{TARGET:} \texttt{MUSLIMS} \\
\hline 
 \textbf{HS:} \textit{ men are more smart than women }\\
\textbf{CN:} \textit{ It is about time women are given a chance to prove that they are intelligent as well. }\\
\textbf{HS$_{pe}$:}  Men are smarter than women, who should be excluded from education. \\
\textbf{CN$_{pe}$:}  This is not true: it is about time women are given a chance to prove that they are intelligent as well... and it is sad that they still need to prove it.  \\
\textbf{TARGET:} \texttt{WOMEN} \\  
    \hline
\textbf{HS:} \textit{Gays are just a bunch of disgusting animals trying to ruin other animals' lives.} \\
\textbf{CN:} \textit{Being gay is not a problem, it is about orientation.} \\
\hline
\textbf{HS$_{pe}$:} Gays are just a bunch of disgusting animals trying to ruin other people's lives. \\
\textbf{CN$_{pe}$:} Being gay is not a problem, being homophobic is. \\
\textbf{TARGET:} \texttt{LGBT+} \\
\hline 
\textbf{HS:} \textit{ Jews are responsible for the 9/11 attacks. }\\
\textbf{CN:} \textit{ I don't think this idea is true: since the 9/11 attacks were committed by Muslims, it doesn't seem fair to blame Jews. }\\
\textbf{HS$_{pe}$:}  Jews are responsible for the 9/11 attacks. \\
\textbf{CN$_{pe}$:}  I don't think this idea is true: since the 9/11 attacks were committed by al Qaeda extremists it doesn't seem fair to blame Jews. \\
\textbf{TARGET:} \texttt{JEWS} \\
\hline 
  \end{tabular}
  \caption{Examples of HS/CN pairs before and after post-editing with assigned target labels.}
 \label{tab:post-edit-example}
\end{table*}

\end{document}